\documentclass[10pt,journal,compsoc]{IEEEtran}
\ifCLASSOPTIONcompsoc
  \usepackage[nocompress]{cite}
\else
  \usepackage{cite}
\fi
\ifCLASSINFOpdf
  \usepackage[pdftex]{graphicx}
  \DeclareGraphicsExtensions{.pdf,.jpeg,.png}
\else
\fi
\usepackage{amsmath, amssymb}
\usepackage{amsthm}
\usepackage{array}
\usepackage{tabularx}
\usepackage{multirow}
\usepackage[flushleft]{threeparttable}
\usepackage{algorithmic}
\ifCLASSOPTIONcompsoc
  \usepackage[caption=false,font=footnotesize,labelfont=sf,textfont=sf]{subfig}
\else
  \usepackage[caption=false,font=footnotesize]{subfig}
\fi
\usepackage{fixltx2e}

\usepackage{stfloats}
\usepackage{url}
\usepackage{color}

\newcommand{\bx}{\mathbf{x}}
\newcommand{\bX}{\mathbf{X}}
\newcommand{\bU}{\mathbf{U}}
\newcommand{\bD}{\mathbf{\Sigma}}
\newcommand{\bV}{\mathbf{V}}
\newcommand{\bQ}{\mathbf{Q}}
\newcommand{\bR}{\mathbf{R}}
\newcommand{\bPi}{\mathbf{\Pi}}

\begin{document}
%
\title{Sparse-TDA: Sparse Realization of Topological Data Analysis for Multi-Way Classification}
%
%
%
%

\author{Wei~Guo,~\IEEEmembership{Student~Member,~IEEE,}
        Krithika~Manohar,
        Steven~L.~Brunton
        and~Ashis~G.~Banerjee,~\IEEEmembership{Member,~IEEE}
\IEEEcompsocitemizethanks{\IEEEcompsocthanksitem W. Guo is with the Department
of Industrial \& Systems Engineering, University of Washington, Seattle,
WA, 98195. E-mail: weig@uw.edu.
\IEEEcompsocthanksitem K. Manohar is with the Department
of Applied Mathematics, University of Washington, Seattle,
WA, 98195. E-mail: kmanohar@uw.edu.
\IEEEcompsocthanksitem S. L. Brunton is with the Department
of Mechanical Engineering, University of Washington, Seattle,
WA, 98195. E-mail: sbrunton@uw.edu.
\IEEEcompsocthanksitem A. G. Banerjee is with the Department
of Industrial \& Systems Engineering and Department of Mechanical Engineering, University of Washington, Seattle,
WA, 98195. E-mail: ashisb@uw.edu.}}

%
%

\markboth{IEEE Transactions on Knowledge and Data Engineering}%
{Guo \MakeLowercase{\textit{et al.}}: Sparse-TDA}
%



\IEEEtitleabstractindextext{%
\begin{abstract}
Topological data analysis (TDA) has emerged as one of the most promising techniques to reconstruct the unknown shapes of high-dimensional spaces from observed data samples. TDA, thus, yields key shape descriptors in the form of persistent topological features that can be used for any supervised or unsupervised learning task, including multi-way classification. Sparse sampling, on the other hand, provides a highly efficient technique to reconstruct signals in the spatial-temporal domain from just a few carefully-chosen samples. Here, we present a new method, referred to as the Sparse-TDA algorithm, that combines favorable aspects of the two techniques. This combination is realized by selecting an optimal set of sparse pixel samples from the persistent features generated by a vector-based TDA algorithm. These sparse samples are selected from a low-rank matrix representation of persistent features using QR pivoting. We show that the Sparse-TDA method demonstrates promising performance on three benchmark problems related to human posture recognition and image texture classification.    
\end{abstract}

}

\maketitle

\IEEEdisplaynontitleabstractindextext

%
\IEEEpeerreviewmaketitle

\IEEEraisesectionheading{\section{Introduction}\label{sec:introduction}}

%
%
%
%
\IEEEPARstart{M}{ulti}-way or multi-class classification, where the goal is to correctly predict one out of $K$ classes for any data sample, poses one of the most challenging problems in supervised learning. However, a large number of real-world sensing problems in a variety of domains such as computer vision, robotics and remote diagnostics, do consist of multiple classes. Examples include human face recognition for surveillance, object detection for mobile robot navigation, and critical equipment condition monitoring for preventive maintenance. The number of classes in these problems often exceeds ten and sometimes goes up to a hundred depending on the complexity of the sensed system or environment and the number and types of sensor modalities. 

While a whole host of techniques such as artificial neural networks, decision trees, na\"{i}ve Bayes, nearest neighbors, and support vector machines (SVMs) have been successfully applied for binary classification problems, extensions of these techniques have had mixed success in addressing multi-way classification problems with more than a few classes. Other approaches involving hierarchical classification or transformation to binary classification have not been particularly successful either. The success rates diminish further in the absence of a large number of data samples for each of the labeled classes. The primary reason is that all of these methods encounter difficulties in selecting the right set of distinguishing features among the different classes.  

Recent research has started investigating completely new techniques for multi-way classification that attempt to better understand the structure of the underlying high-dimensional sample space. One such class of techniques is topological data analysis, or TDA in short. TDA represents the unknown sample space in the form of persistent shape descriptors that are coordinate free and deformation invariant. Thus, the descriptors define topological features and yield insights regarding suitable feature selection. 

Another critical tool facilitating multi-way classification is the feature-driven sparse sampling of high-dimensional data. Observations are typically sparse in a transform basis of the informative features, so that samples can be optimally chosen to enhance the discriminating features in the data. This sparsity permits heavily subsampled inputs for downstream classifiers, which drastically reduces the burdens of sample acquisition, processing and storage without sacrificing performance. \looseness=-1

Here, we bring together the two research areas of TDA and sparse sampling in the context of multi-way classification. In particular, we leverage QR pivoting-based sparse sampling for optimal feature selection once the topological features are extracted using a state-of-the-art TDA method. We test our method on three challenging data sets pertaining to 3D meshes of synthetic and real human postures and textured images, respectively. We call our new method the Sparse-TDA algorithm. We show that it achieves comparable accuracy as the kernel TDA method with substantially lower training times, and better accuracy with comparable or lower training times than widely-used L1-regularized classifiers. Thus, our method opens up a new direction in making online multi-way classification practically feasible.  \looseness=-1

\section{Related Work}
\label{sec:review}
Over the past decade or so, an increasing interest in utilizing tools from algebraic topology to extract insights from high dimensional data has given rise to the field of TDA. The successful applications of TDA have spanned a large number of areas, ranging from computer vision \cite{Reininghaus15} 
to medical imaging \cite{gao2013segmenting}, biochemistry \cite{gameiro2015topological}, neuroscience \cite{chung2009persistence} and materials science \cite{hiraoka2016hierarchical}. A predominant tool in TDA is persistent homology, which tracks the evolution of the topological features in a multi-scale manner to avoid information loss \cite{edelsbrunner2002topological,zomorodian2005computing}. The multi-scale information is summarized by the persistence diagram (PD), a multiset of points in $\mathbb{R}^2$ that encodes the lifetime (i.e., persistence) of the features. 

More recently, researchers have started utilizing TDA for machine learning problems. Pachauri et al. \cite{pachauri2011topology} first computed a Gaussian kernel to estimate the density of points on a regular grid for each rasterized PD, and fed the discrete density estimation as a vector into an SVM classifier without any feature selection. However, their method did not establish the stability of the kernel-induced vector representation. Reininghaus et al. \cite{Reininghaus15} then designed a stable multi-scale kernel for PDs motivated by scale-space theory as will be described in the next Section. Experiments on three benchmark data sets showed that this method greatly outperformed an alternative approach based on persistence landscape \cite{bubenik2015statistical}, a popular statistical treatment of TDA. Similar to this work, Kusano et al. \cite{kusano2016persistence} proposed a stable persistence weighted Gaussian kernel, allowing one to control the effect of persistence. However, the computational complexity of both the kernel-based methods for calculating the Gram matrix is $O(m^2n^2)$ if there are $n$ PDs for training and the PDs contain at most $m$ points, which can be quite expensive for many practical applications. 

To enable large-scale computations with PDs, recent methods have mapped each PD to a stable vector to allow direct use of vector-based learning methods. For example, Adams et al. \cite{Adams16} constructed vectors by discretizing the weighted sum of probability distributions centered at each point in transformed PDs. Carri{\`e}re et al. \cite{carriere2015stable} rearranged the entries of the distance matrix between points in a PD and Bonis et al. \cite{bonis2016persistence} adopted a pooling scheme to construct the vectors. 

Sparse optimized sampling of vectorized PDs can provide a further reduction for improved classifier training performance, by leveraging an initial low-rank feature transformation such as principal components analysis (PCA).
In the context of image classification using linear discriminant analysis, Brunton et al.~\cite{Brunton2016siap} use convex $\ell_1$ optimization to identify sparse pixel locations that map into the discriminating subspaces in PCA coordinates.
Recent advances in model order reduction employ fast matrix pivoting schemes to sample PCA libraries for sparse classification of dynamical regimes in physical systems~\cite{Sargsyan2015pre,Manohar2016jfs}.

In this work, we employ the vector representation from \cite{Adams16} and integrate with a sparse sampling method using QR pivots to identify discriminative features in the presence of noisy and redundant information to further improve classifier training time and sometimes prediction accuracy.

\section{Sparse-TDA Method}
\label{sec:sparse-TDA}
We now introduce a vector representation of a PD, termed a persistence image (PI), presented in \cite{Adams16}. Since our Sparse-TDA method will combine PI-based TDA with sparse sample selection, we first summarize the sparse sampling method before describing the combination.  

\subsection{Optimized Sparse Sample Selection}
Vectorized PIs sparsely encode topological structure within a few key pixel locations containing nonzero entries. 
Sampling these PIs at critical pixel locations is often sufficient for training downstream classifiers at a fraction of the runtime required for full PIs.
To determine these PI indices, we use a pixel sampling method based on powerful low-rank matrix approximations. First, we arrange the PI vectors from all the training classes into columns of a matrix $\bX$ and compute its truncated singular value decomposition to obtain the dominant PI variation patterns (principal components) $\bU_r$
\begin{equation}
\label{eqn_svd}
\bX \approx \bU_r\bD_r\bV_r^T.
\end{equation}
The SVD truncation parameter $r$ determines the number of pixel samples and is chosen according to the optimal singular value threshold~\cite{GavishDonoho2014}. We then discretely sample the PI principal components using the pivoted QR factorization, an efficient greedy alternative to expensive convex optimization methods. QR pivoting is the workhorse behind discrete sampling for underdetermined least squares problems~\cite{Businger1965}, polynomial interpolation~\cite{Sommariva2009cma}, and more recently, model order reduction~\cite{Drmac15} and sensor placement~\cite{Manohar2017ieee}. The pivoting procedure optimizes a row permutation $\bPi$ of the principal components that is numerically well-conditioned by factoring $\bU_r^T$ into unitary and upper-triangular matrices $\bQ$ and $\bR$ \looseness=-1
\begin{equation}
\bU_r^T\bPi^T = \bQ\bR.
\end{equation}
The final step converts a given PI, $\bx$, into a sparsely sampled PI, ${\tilde\bx = \bPi_r\bx}$, where the first $r$ permutation indices correspond to the selected pixel locations.

\subsection{Combining Sparse Sample Selection with Persistence Images}
\label{sec:our_method}
Let $\mathcal{D} =\{D_i\ |\ i=1, \ldots, n\}$ be a training set of PDs. To construct a PI from a given PD $D_i$ \cite{Adams16}, $D_i$ is first transformed from birth-death coordinates to birth-persistence coordinates. Let $T: \mathbb{R}^2 \rightarrow \mathbb{R}^2$ be the linear transformation, 
\begin{equation}
T(x, y) = (x, y-x).
\end{equation}
A persistence surface $\rho_{D_i}: \mathbb{R}^2 \rightarrow \mathbb{R}$ on $T(D_i)$ is defined by
\begin{equation}
\rho_{D_i}(z) = \sum_{u \in T(D_i)} f(u)g_u(z)
\end{equation}
where $f: \mathbb{R}^2 \rightarrow \mathbb{R}$ is a non-negative weighting function that is zero along the horizontal axis, continuous, and piecewise differentiable; $g_u: \mathbb{R}^2 \rightarrow \mathbb{R}$ is a probability function with mean $u = (u_x, u_y) \in \mathbb{R}^2$ and variance $\sigma^2$.

In our experiments, the linear weighting (LW) function is 
\begin{equation}
f(u) =\frac{u_y}{u_y^*},
\end{equation}
where $u_y^* = \max_{i=1, \ldots, n}\max_{u \in D_i} u_y$. The form of the nonlinear weighting (NW) function is inspired by the weighting function used in \cite{kusano2016persistence} and chosen as
\begin{equation}
f(u) = \arctan(c u_y).
\end{equation}
where $c = ({\rm median}_{i=1, \ldots, n}{\rm median}_{u\in D_i} u_y)^{-1}$. We choose $g_u$ to be the Gaussian distribution, i.e.,
\begin{equation}
g_u(z) = \frac{1}{2\pi\sigma^2}e^{-[(z_x-u_x)^2+(z_y-u_y)^2]/2\sigma^2}.
\end{equation}
where $z = (z_x,z_y)$. Then the PI, a matrix of pixel values, is obtained by calculating the integral of $\rho_{D_i}$ on each grid box from discretization,
\begin{equation}
I(\rho_{D_i}) = \iint \rho_{D_i}(z_x, z_y)dz_x dz_y.
\end{equation}
PI has also been proven to be 1-Wasserstein stable. 
Assume that the number of desired features (i.e., pixel samples) is $s$. Applying the sparse sampling method on $\mathbf{X}$, we obtain the row indices of $s$ optimal pixel locations and the sparsely sampled PIs $\tilde{\mathbf{x}}_1, \ldots, \tilde{\mathbf{x}}_n \in \mathbb{R}^s$ for the downstream classifiers. \looseness=-1 

\section{Results}
\label{sec:results}
We now discuss the performance of our Sparse-TDA method on three benchmark computer vision data sets. The data sets are explained first, followed by illustrations of the selected features, and quantitative comparisons of our method with the L1-SVM feature selection method using the same PIs and the multi-scale kernel TDA method. The illustrations and comparison results show the usefulness of the method on challenging multi-way classification problems.    

\subsection{Data Sets}
For shape classification, SHREC'14 synthetic and real data sets are used, given in the format of triangulated 3D meshes \cite{pickup1414}. The synthetic set contains meshes from five males, five females and five children in 20 different poses, while the real set consists of 20 males and 20 females in 10 different poses. 

For texture recognition, we use the \textsf{Outex\_TC\_00000} data set \cite{Ojala02}. This data set contains 480 images equally categorized into 24 classes and provides 100 predefined 50/50 training/testing splits. During preprocessing, we downsample the original images to $32 \times 32$ pixel images as done in the multi-scale kernel TDA method. 

\begin{figure*}[!t]
\centering
\subfloat[SHREC'14 synthetic postures]{\includegraphics[width=2.25in]{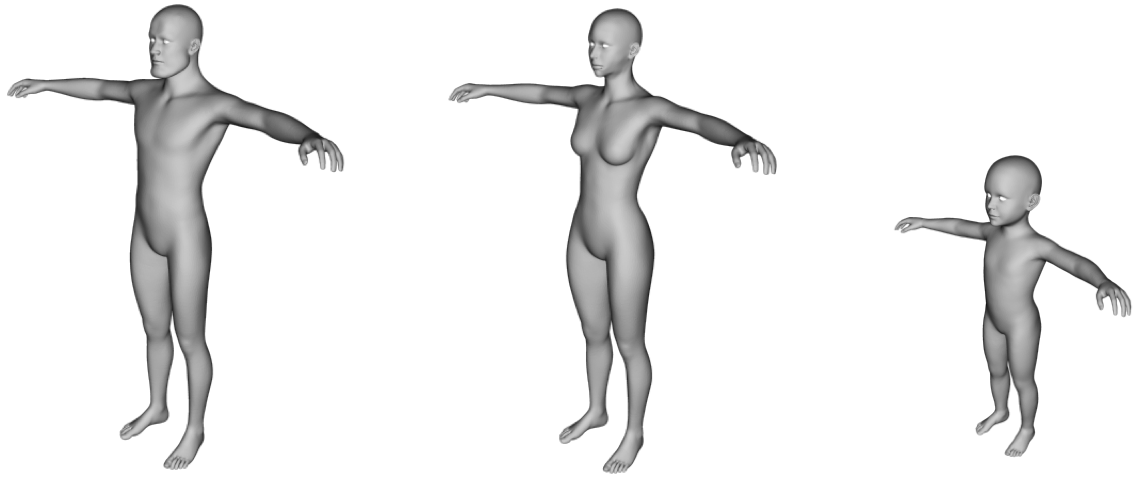}%
\label{fig_visualization_syn}}
\hfil
\subfloat[SHREC'14 real postures]{\includegraphics[width=2.25in]{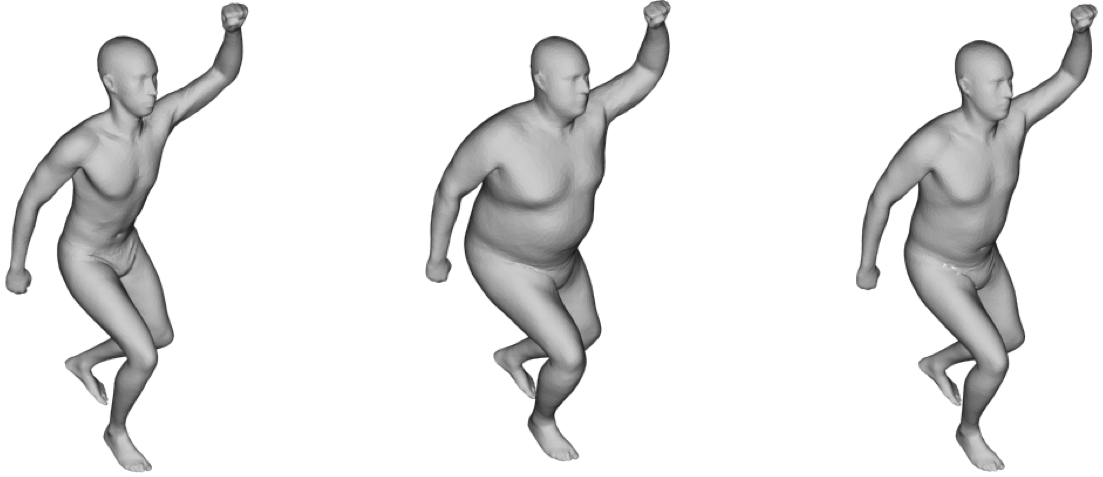}%
\label{fig_visualization_real}}
\hfil
\subfloat[OuTeX texture images]{\includegraphics[width=2.25in]{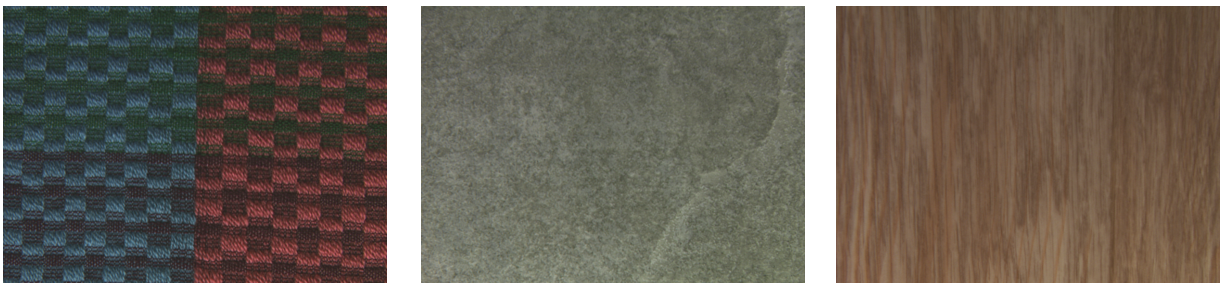}%
\label{fig_visualization_outex}}
\vfil
\subfloat[SHREC'14 synthetic PIs using LW]{\includegraphics[width=2.25in]{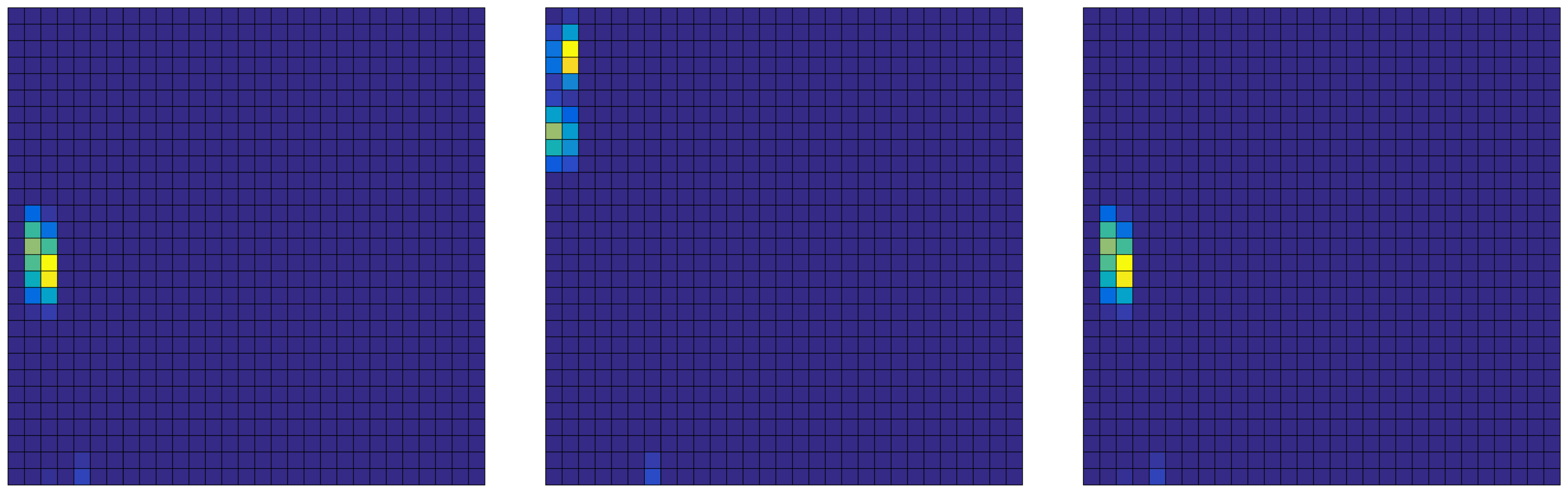}%
\label{fig_PI_syn_linear}}
\hfil
\subfloat[SHREC'14 real PIs using LW]{\includegraphics[width=2.25in]{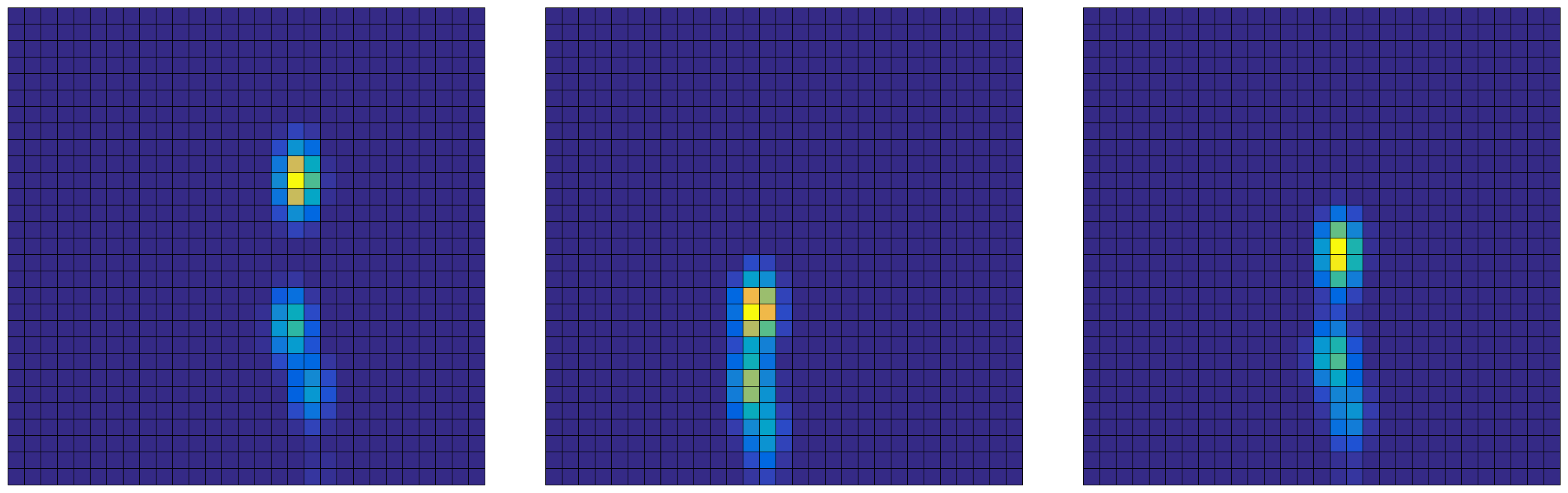}%
\label{fig_PI_real_linear}}
\hfil
\subfloat[OuTeX texture PIs using LW]{\includegraphics[width=2.25in]{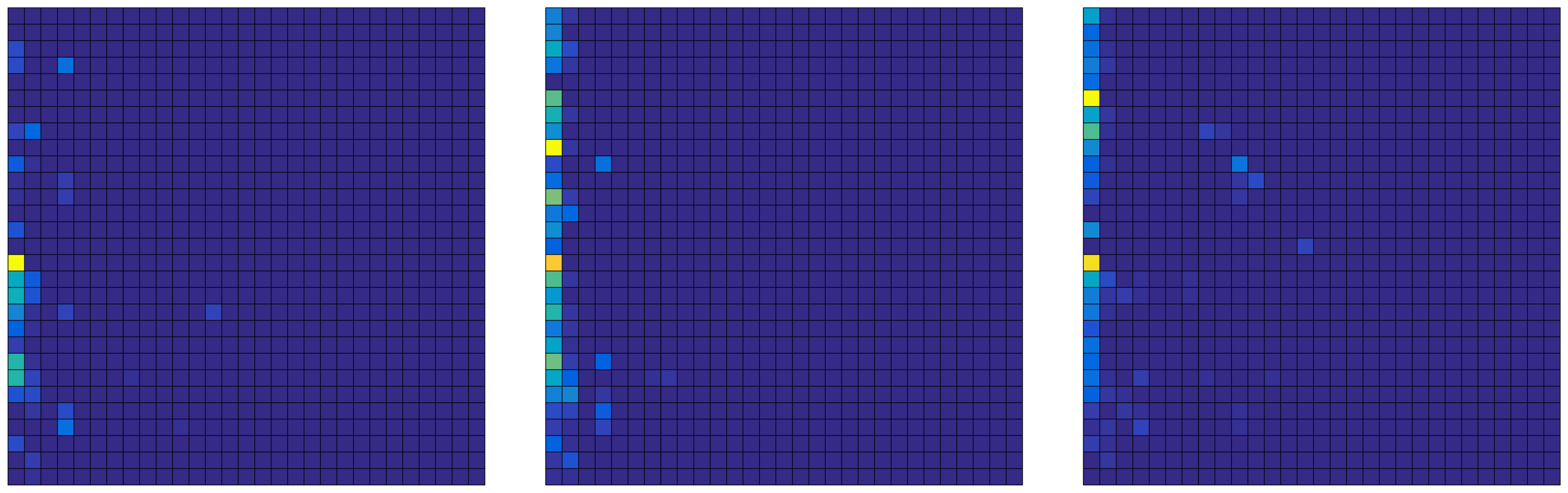}%
\label{fig_PI_outex_linear}}
\vfil
\subfloat[SHREC'14 synthetic PIs using NW]{\includegraphics[width=2.25in]{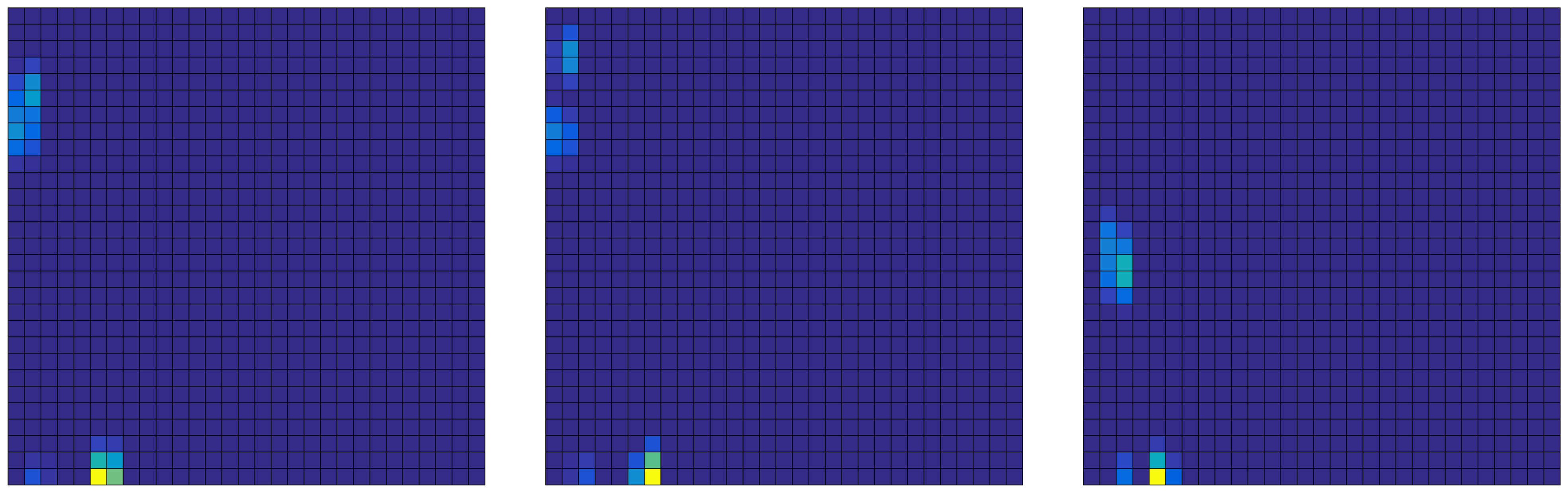}%
\label{fig_PI_syn_nonlinear}}
\hfil
\subfloat[SHREC'14 real PIs using NW]{\includegraphics[width=2.25in]{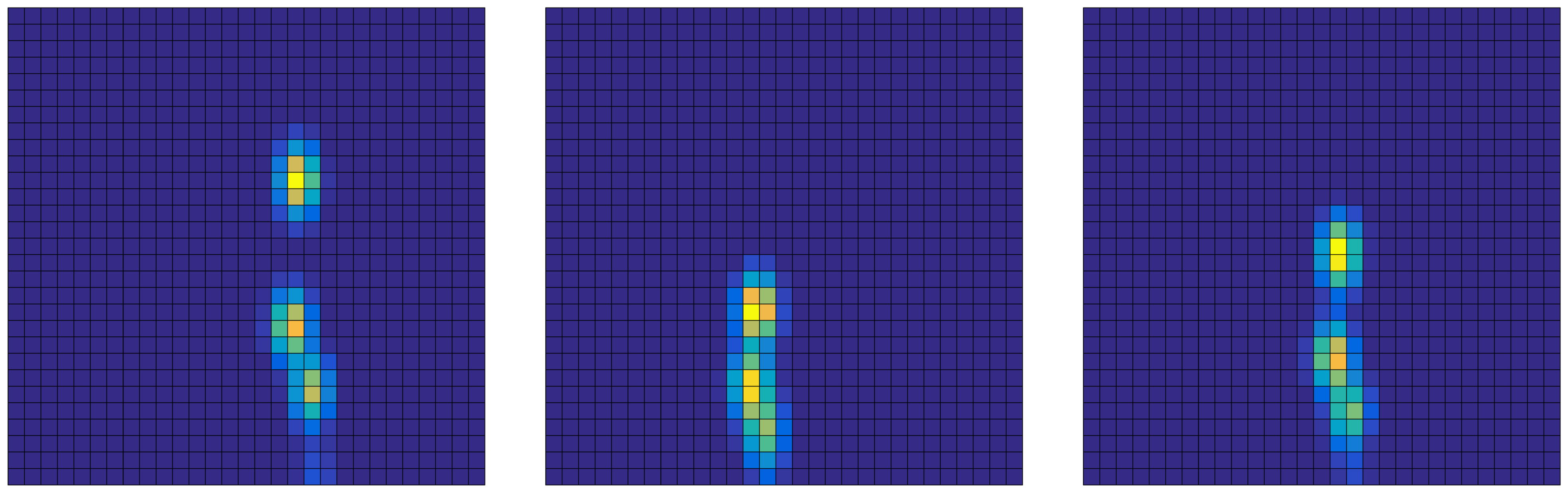}%
\label{fig_PI_real_nonlinear}}
\hfil
\subfloat[OuTeX texture PIs using NW]{\includegraphics[width=2.25in]{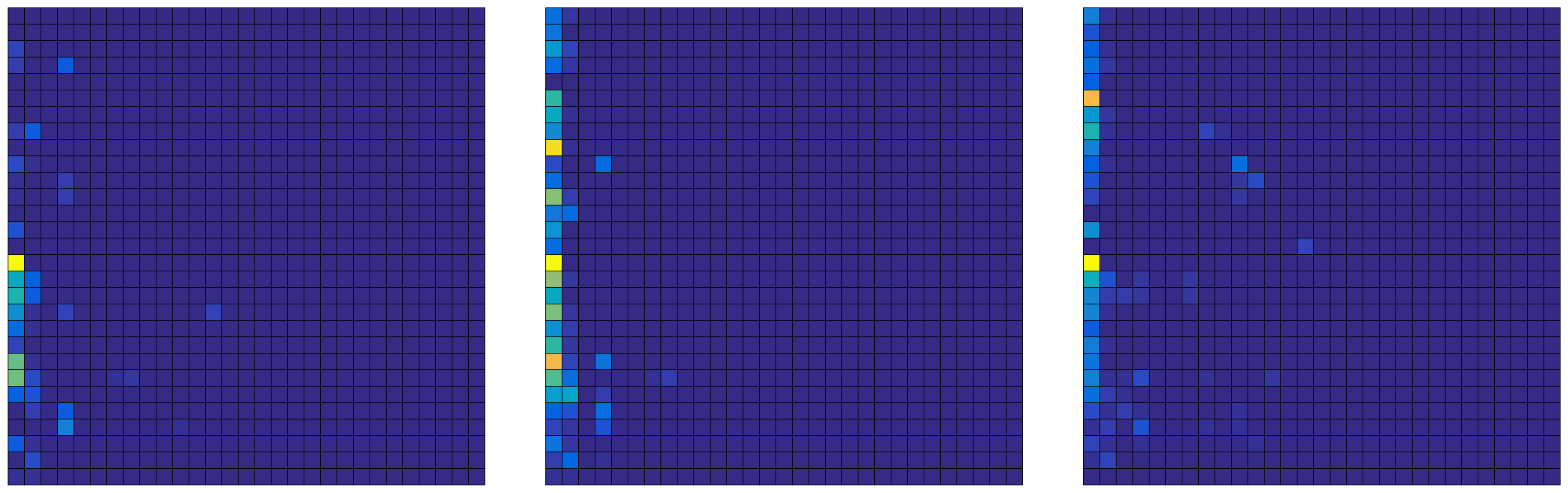}%
\label{fig_PI_outex_nonlinear}}
\caption{Representative classes and corresponding persistence images (PIs) for various benchmark data sets using two Sparse-TDA variants.}
\label{fig_data_images}
\end{figure*}

\subsection{Feature Selection}
We first follow the same procedure performed in the multi-scale kernel TDA method to obtain the PDs. For SHREC'14 data sets, we compute the heat kernel signature \cite{sun2009concise} on the surface mesh of each object and then compute the 1-dimensional PDs using Dipha\footnote{https://github.com/DIPHA/dipha}. For the OuTeX data set, we take the sign component of the completed local binary pattern operator \cite{guo2010completed} as the descriptor function. Then we generate the 0-dimensional PDs from the filtration of its rotation-invariant version with $P=8$ neighbors and radius $R=1$. \looseness=-1
 
\begin{figure*}[!t]
\centering
\subfloat[SHREC'14 Synthetic. Training size: 210. Linear weighting: $r_o = 100\sim103$. Nonlinear weighting: $r_o = 99\sim104$.]{
\includegraphics[width=2.35in]{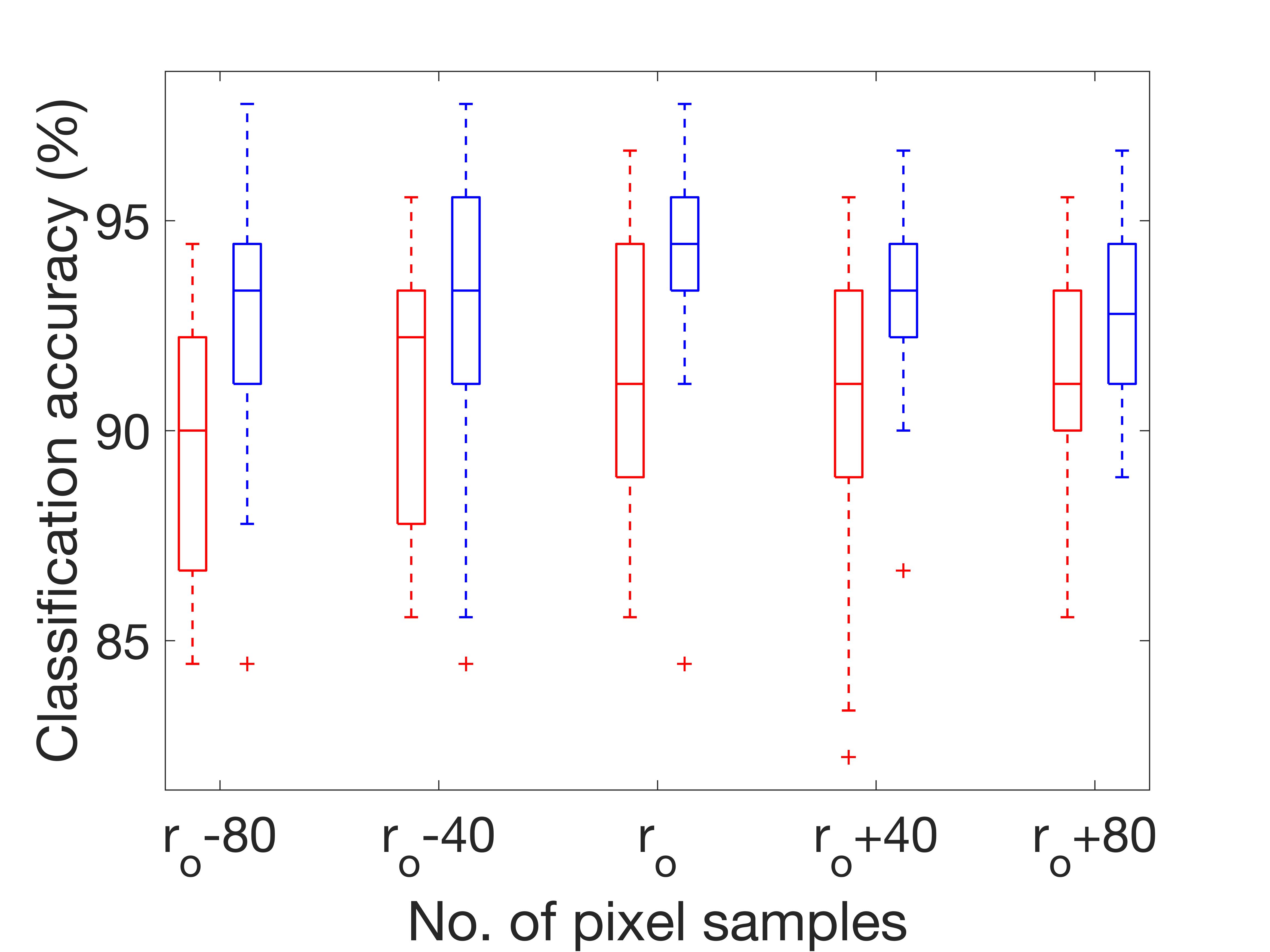}
\includegraphics[width=2.35in]{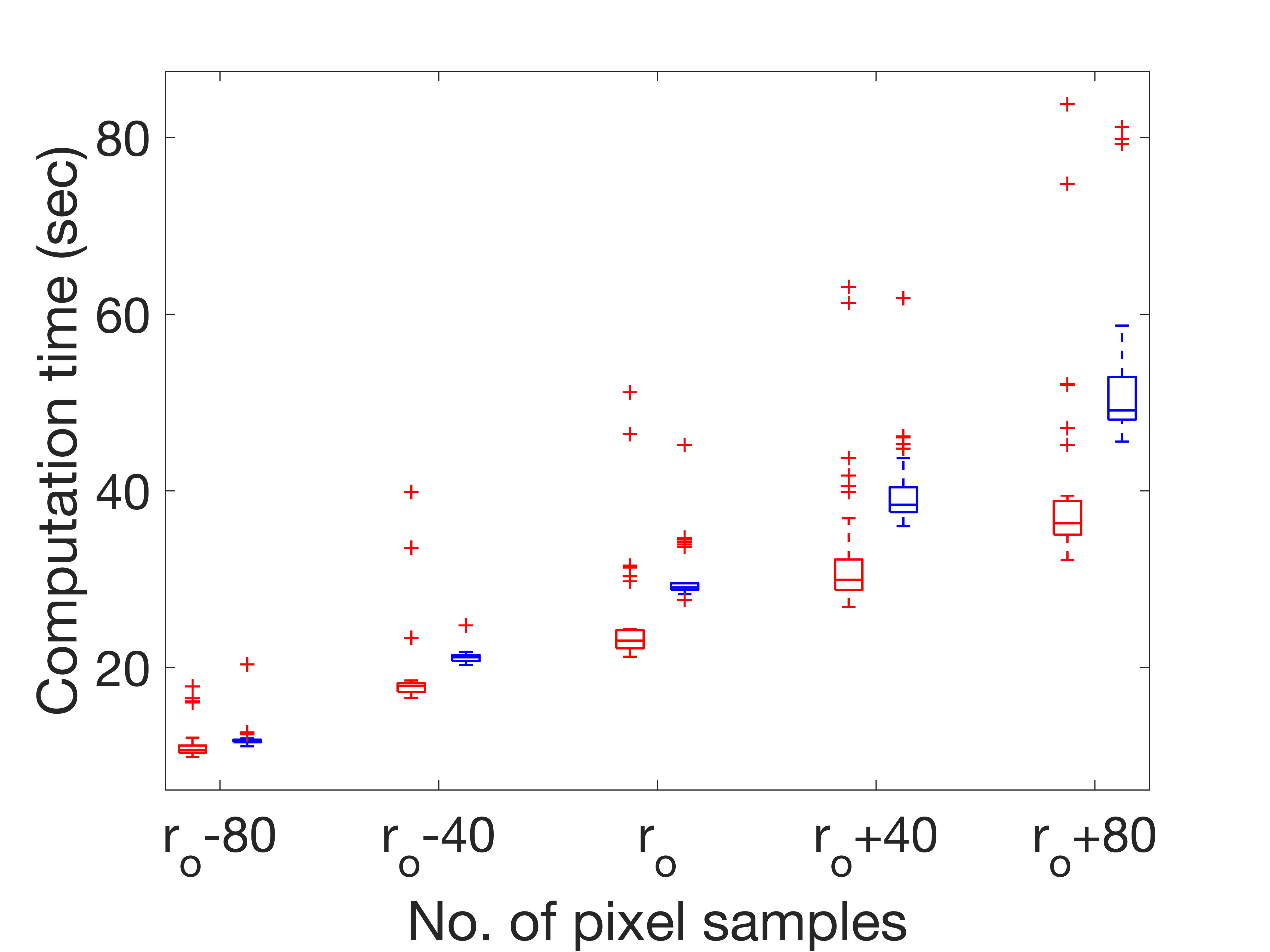}
\includegraphics[width=2.35in]{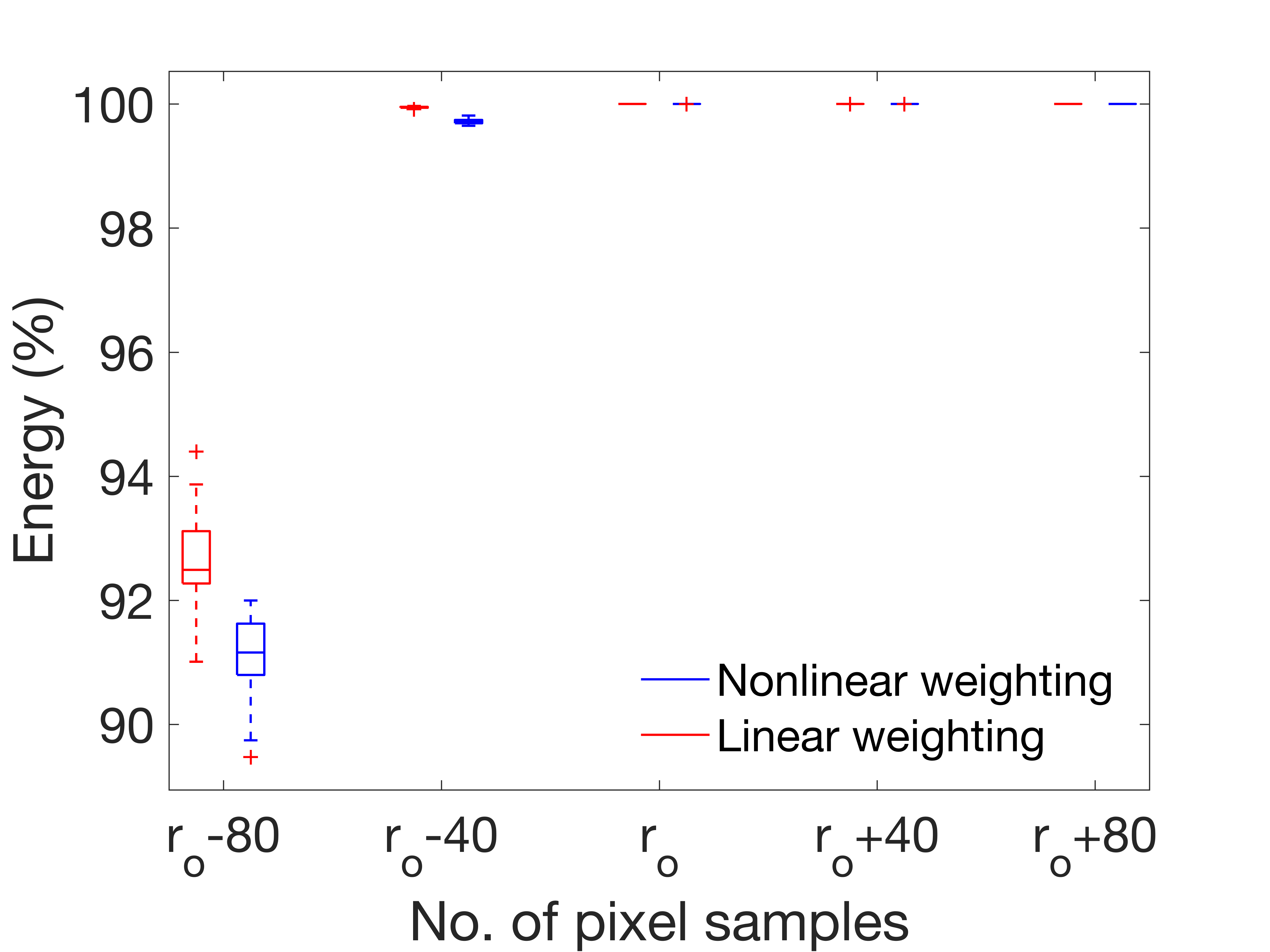}
\label{fig_SHREC_syn_multiple_s}}
\vfill
\subfloat[SHREC'14 Real. Training size: 280. Linear weighting: $r_o = 121\sim126$. Nonlinear weighting: $r_o = 122\sim126$.]{
\includegraphics[width=2.35in]{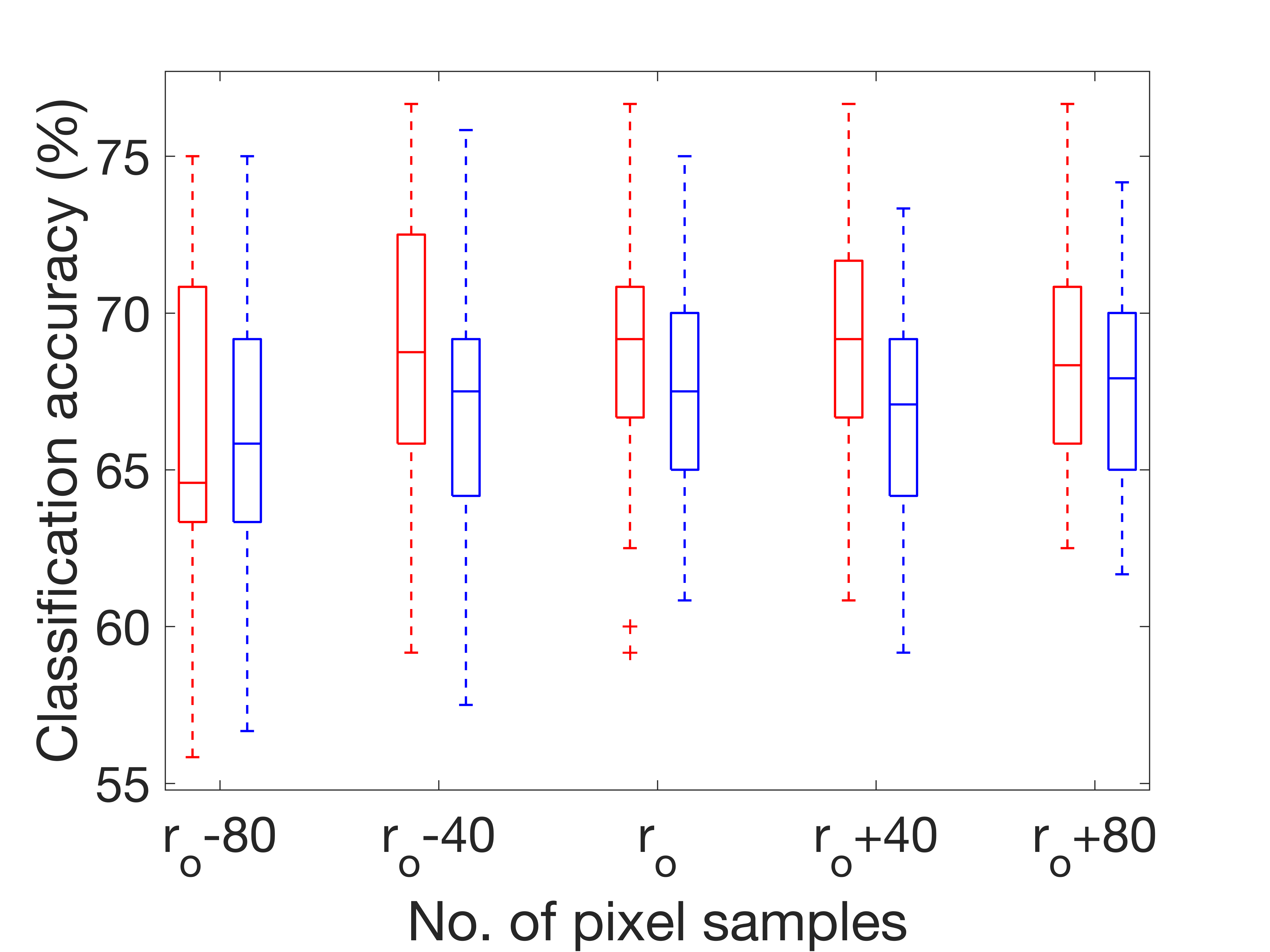}
\includegraphics[width=2.35in]{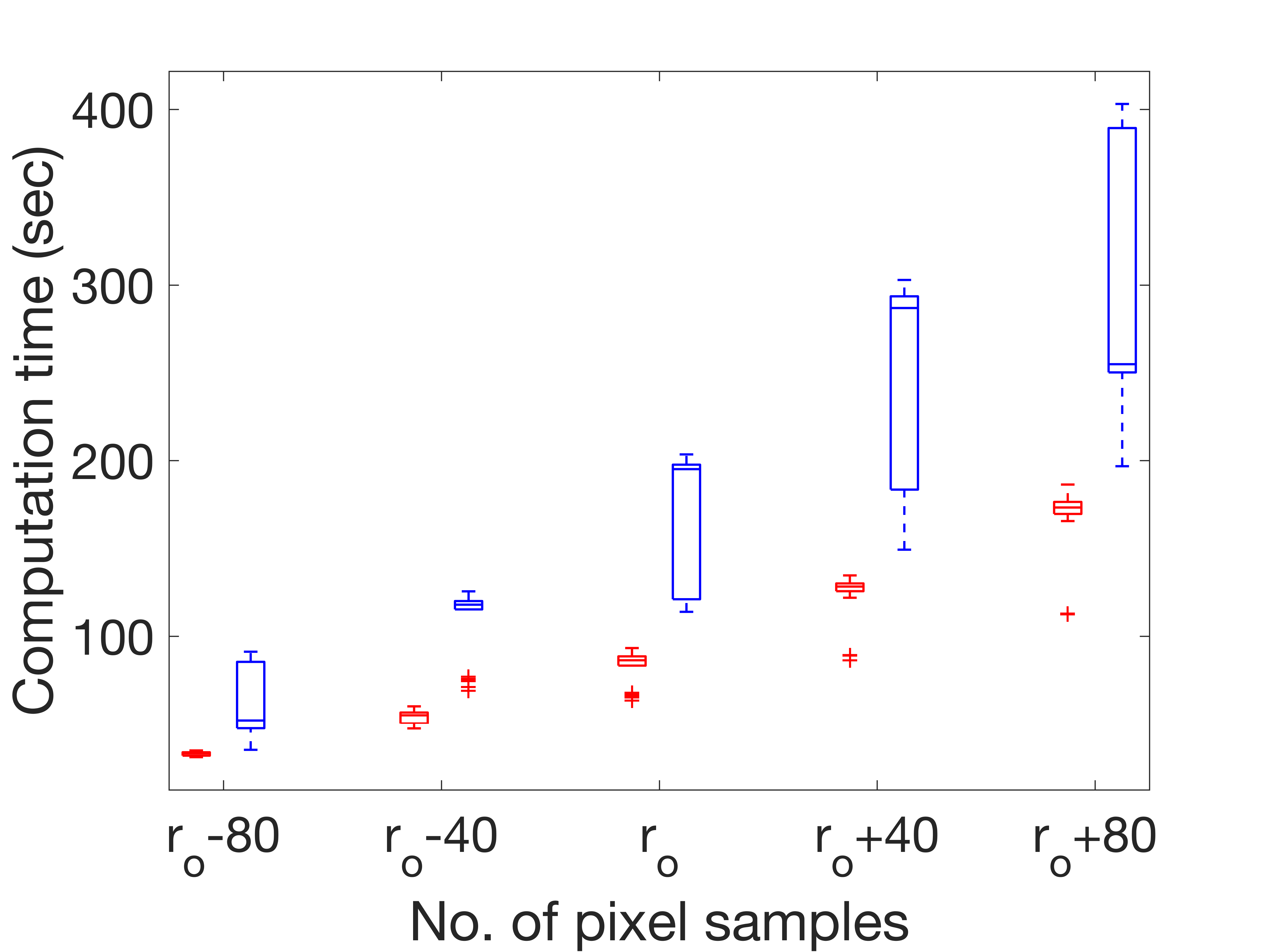}
\includegraphics[width=2.35in]{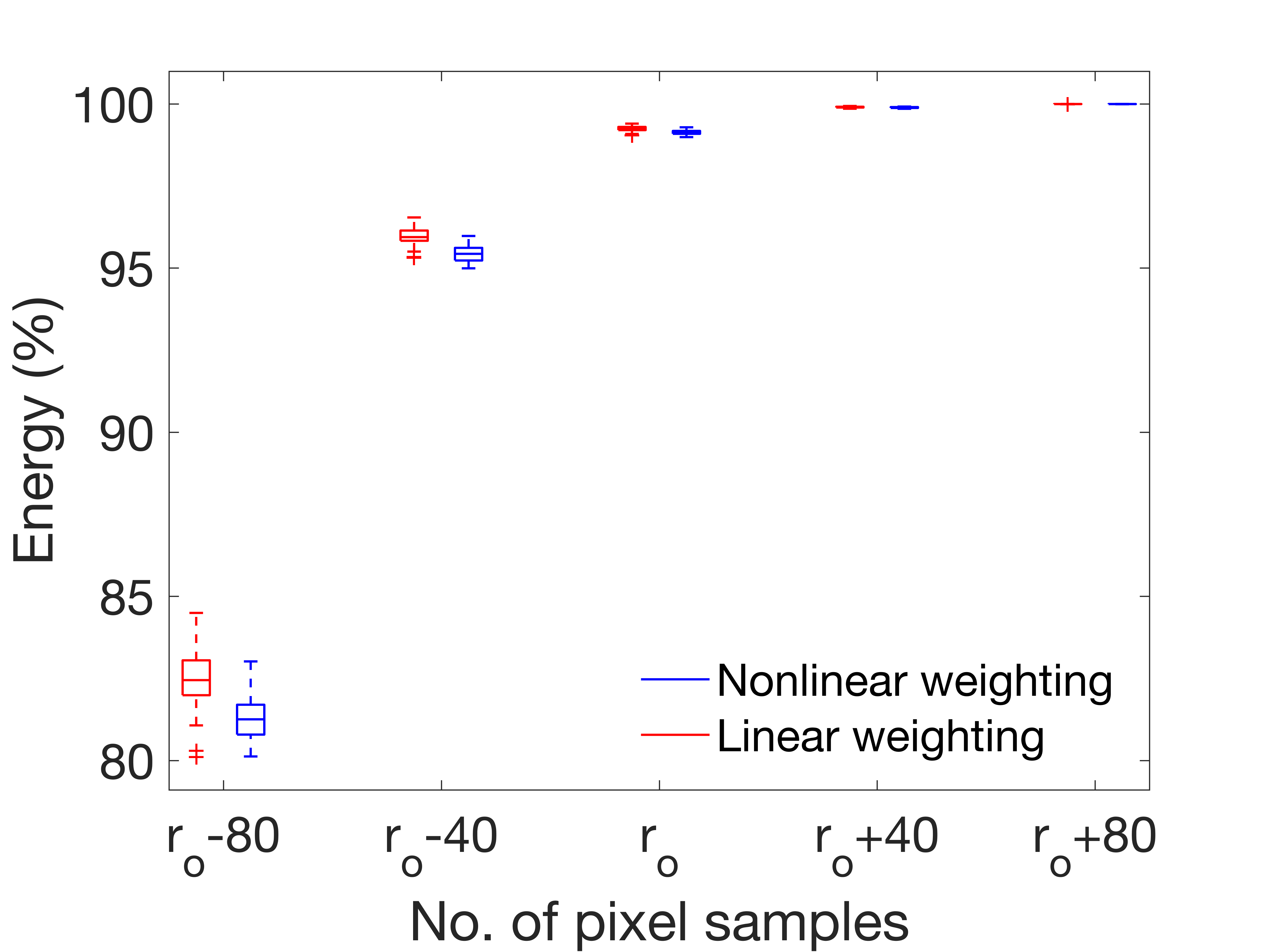}
\label{fig_SHREC_real_multiple_s}}
\vfill
\subfloat[OuTeX Texture. Training size: 336. Linear weighting: $r_o=112\sim 116$. Nonlinear weighting: $r_o=108\sim 112$.]{
\includegraphics[width=2.35in]{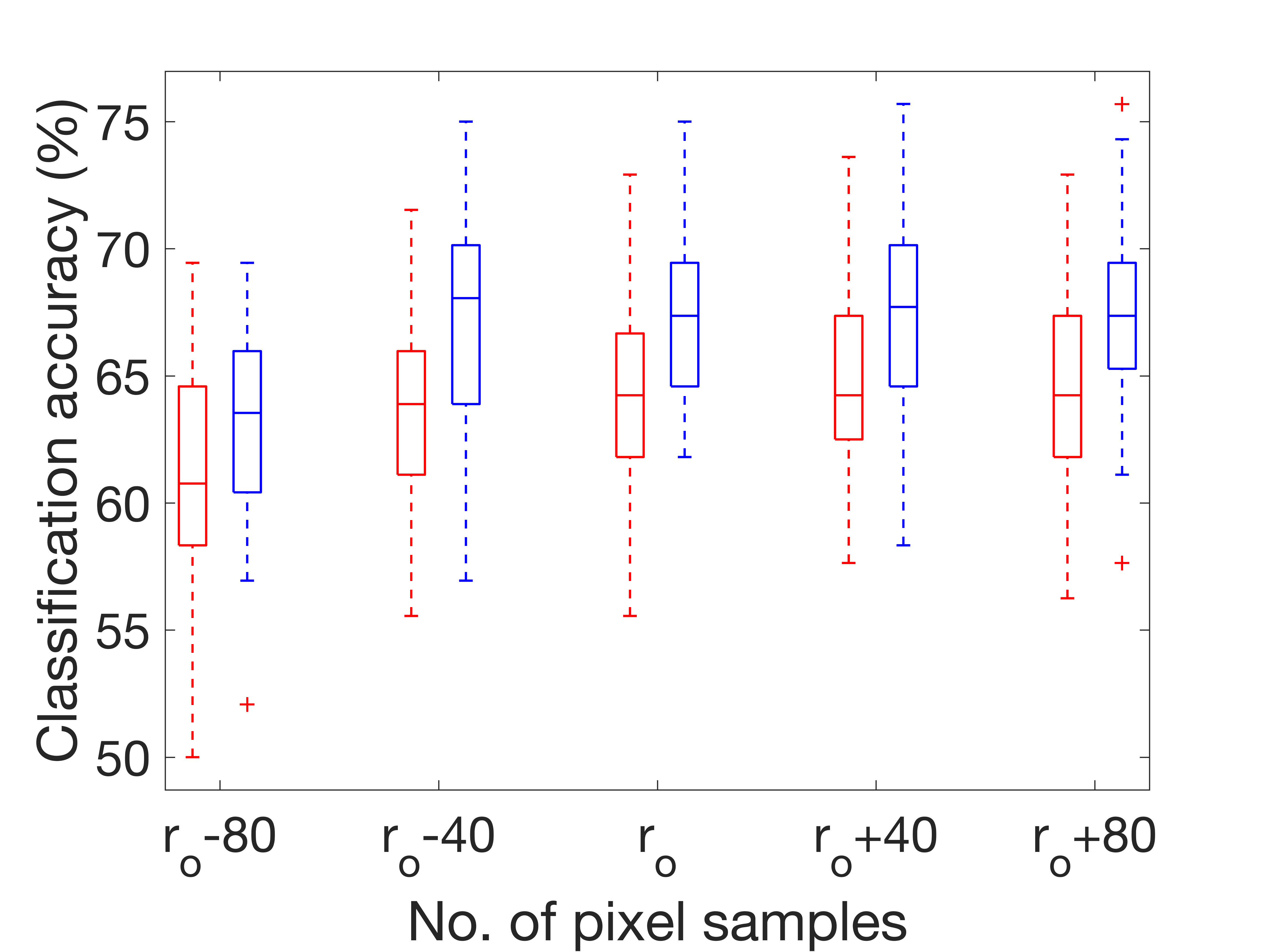}
\includegraphics[width=2.35in]{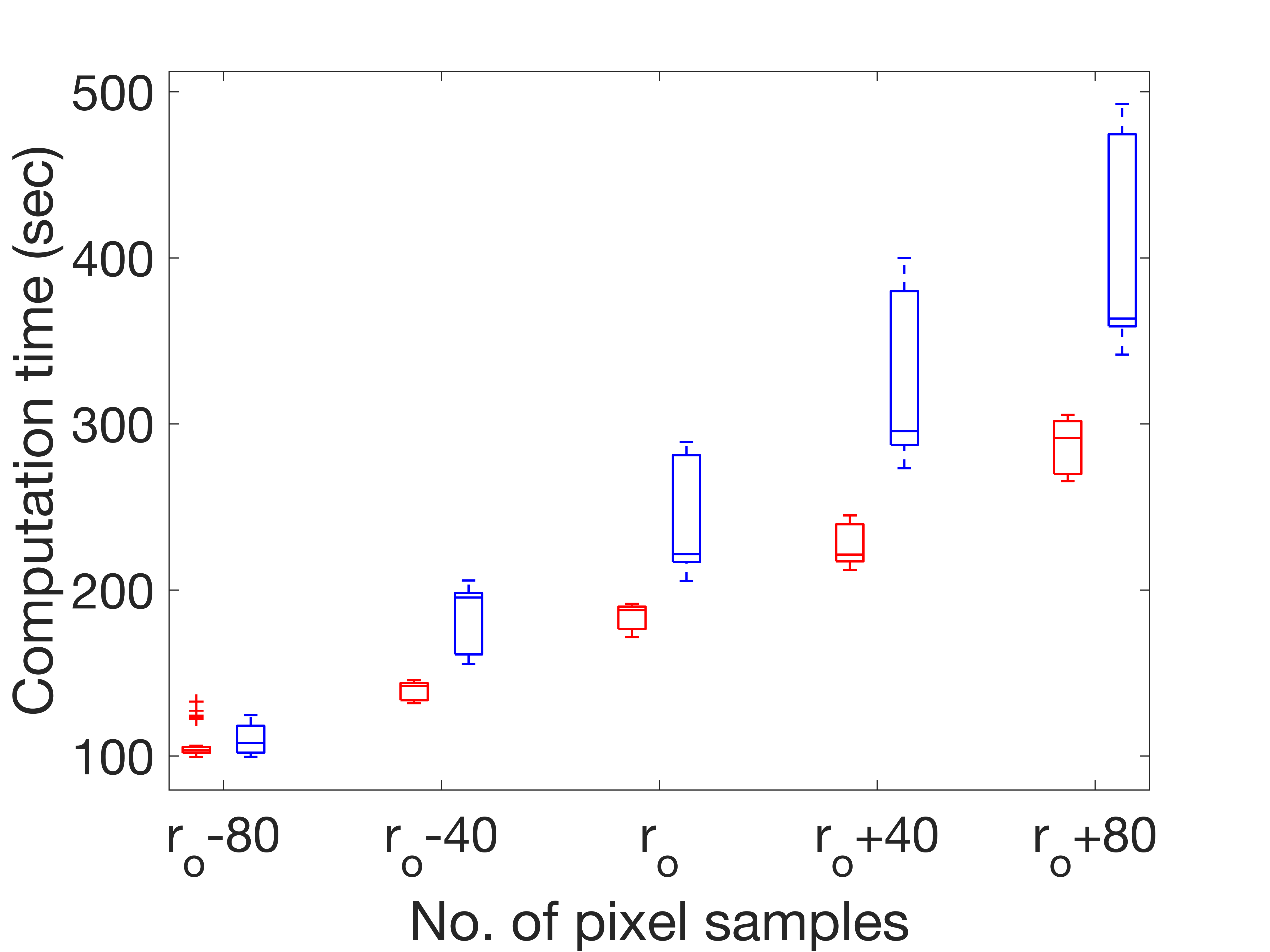}
\includegraphics[width=2.35in]{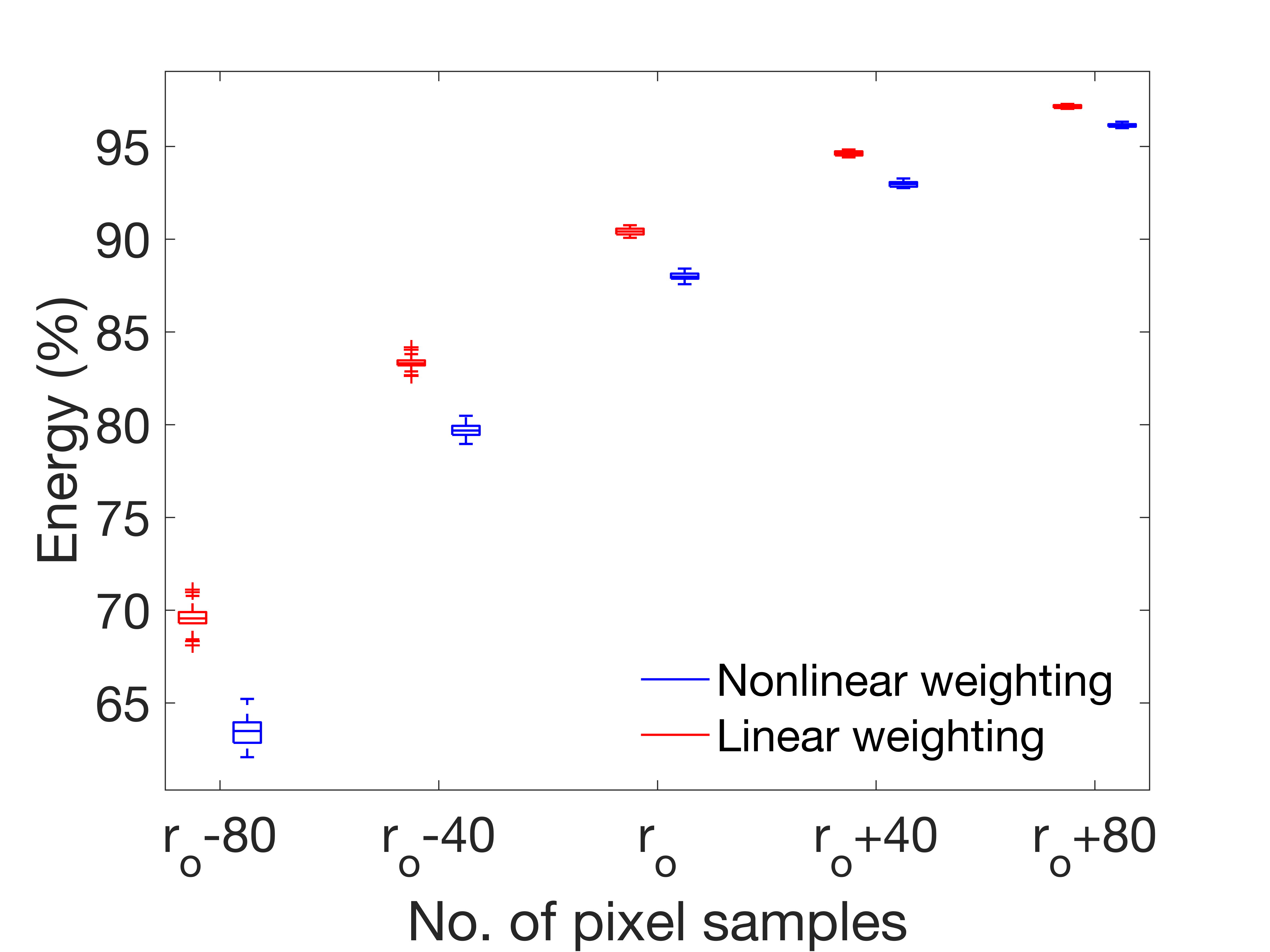}
\label{fig_Outex_multiple_s}}
\caption{Comparison of classification accuracy, training time, and energy among different selections of desired pixel samples for the benchmark data sets using two Sparse-TDA methods. The results are based on 30 runs with a 70/30 training/testing split.}
\label{fig_multiple_s}
\end{figure*}

To generate the PIs, we set the grid resolution to be 30 $\times$ 30 for all three data sets. In fact, the classification accuracy is fairly robust to the choice of resolution \cite{Adams16}. We also set $\sigma$ to be 0.2, 0.0001 and 0.02 for SHREC'14 synthetic, SHREC'14 real and OuTex data sets, respectively. Fig.~\ref{fig_data_images} shows representative PIs for three different classes in all of our benchmark data sets. Noticeable differences are observed among the PIs for each of the three data sets, although the differences are most pronounced for the SHREC'14 synthetic data set, reasonably clear for the SHREC'14 synthetic data set, and less evident for the OuTeX data set. These differences in the pixel values of the PIs form the distinguishing class features from which an optimal set is selected by QR pivots. 
Fig.~\ref{fig_multiple_s} measures the effect of varying the number of pixel samples determined by the SVD truncation parameter in Eq.~\eqref{eqn_svd} on classification accuracy and performance (see below for detailed settings). As expected, classifier training time increases with additional samples. The accuracy, however, improves until the number of samples equals the optimal SVD truncation parameter $s=r_o$, after which limited additional information is available. Beyond this value, accuracy tapers off, which is consistent with the percentage of PI variance (energy) captured by the truncated SVD. For this reason, $s$ is selected as $r_o$ for our Sparse-TDA method in the following simulations. In the case of the L1-SVM method, a sparse solution is generated by L1 regularization during the training phase of a linear classifier. No feature selection is involved for the kernel TDA method.

\subsection{Classification Performance}
We feed the reduced feature vectors for training into a soft margin C-SVM classifier with a radial basis function (RBF) kernel, implemented in \textsf{LIBSVM}~\cite{chang2011libsvm}, for each data set. The cost factor $C$ and kernel parameter $\gamma$ are tuned based on a grid search using 10-fold cross-validation on the training data. We start a coarse grid search with exponentially growing sequences of $C$ and $\gamma$ first, thereafter proceeding with finer grid searches in the vicinity of the optimal region yielded by the previous grid search. Each grid search includes a total of 50 pairs of $(C,\gamma)$ values which are used to apply the training model to the sparsely sampled PIs
of the test set. For the L1-SVM method, since only the cost factor needs to be trained, it is then tuned 10 times using the same scheme as described above with the implementation in \textsf{LIBLINEAR}~\cite{fan2008liblinear}. Results are reported based on 30 runs for each case with the exception of those presented for the OuTex data set in Tables~\ref{table_accuracy}-\ref{table_time}, which are based on 100 runs.\looseness=-1

Table~\ref{table_accuracy} compares the classification accuracy of both the variants of the L1-SVM and our Sparse-TDA method with the multi-scale kernel TDA method. Note that the two SHREC'14 data sets are partitioned into 70/30 training/testing samples, whereas the OuTeX set is partitioned into 50/50 training/testing samples. The number of samples for each class is approximately the same in all the training sets. 
Consistent with the differences observed in the PIs among the classes, both the variants of our method perform slightly better than the kernel TDA method for the SHREC'14 real data set. On the other hand, our method is marginally worse than the kernel TDA method for both the SHREC'14 synthetic and OuTeX data sets, even though the accuracy increases slightly using nonlinear weighting. Both the L1-SVM variants are, however, inferior to the other methods by varying degrees for all the data sets.

\begin{table}[h]
\renewcommand{\arraystretch}{1.25}
\caption{Comparison of classification accuracy (in \%) with the same training and test data set split as in the original multi-scale kernel TDA article}
\centering
\begin{tabular}{|c|c|c|c|c|}
\hline
\multicolumn{2}{|c|}{\multirow{2}{*}{Method}} & SHREC'14 & SHREC'14 & OuTeX \\
\multicolumn{2}{|c|}{} & Synthetic & Real & Texture\\
\hline 
\multirow{2}{*}{L1-SVM} & LW & $89.6\pm 2.3$ & $63.9\pm 4.9$ & $55.1\pm 3.5$\\
\cline{2-5}
& NW & $92.1\pm 2.5$ & $63.9\pm 4.4$ & $57.4\pm 3.6$ \\
\hline 
\multirow{2}{*}{Sparse-TDA} & LW & $91.5\pm 3.0$ & ${\bf 68.8\pm 4.2}$ & $62.6\pm 2.7$ \\
\cline{2-5}
& NW & $94.0\pm 2.6$ & $67.8\pm 3.6$ & $66.0\pm 2.4$ \\
\hline 
\multicolumn{2}{|c|}{Kernel TDA} & ${\bf 97.8\pm 2.0}$ & $65.3\pm 4.5$ & ${\bf 69.2\pm 2.4}$ \\
\hline
\end{tabular}
\label{table_accuracy}
\end{table}

Table~\ref{table_time} provides a comparison of the three methods in terms of the SVM-based classifier training time as measured on a laptop with a 2.4GHz Intel Core i5 CPU and 4 GB RAM. In the case of the L1-SVM and Sparse-TDA methods, the training time starts from the computation of the PIs. Not surprisingly, both our method variants are usually much faster than kernel TDA as they use smaller sets of selected features. As expected, the reduction in training time is greater with linear weighting than with nonlinear weighting. In fact, the Sparse-TDA method with linear weighting achieves about 46X speed-up for the SHREC'14 synthetic data set and roughly 45X speed-up for the OuTeX data set. However, there is no consistent speed-up for the SHREC'14 real data set owing to the fact that there are only 4-5 points in each PD, rendering the training of the kernel TDA method exceptionally fast. In contrast, there are 38-294 points and 127-299 points in each PD for the SHREC'14 synthetic and OuTeX data sets, respectively. On the other hand, the L1-SVM method is not consistently fast due to the non-differentiability of the L1-regularized form, which leads to more difficulties in solving the optimization problem during training. For example, the training time of the L1-SVM method with linear weighting is more than three times as much as that of our counterpart method.

\begin{figure*}[!t]
\centering
\subfloat{\includegraphics[width=2.35in]{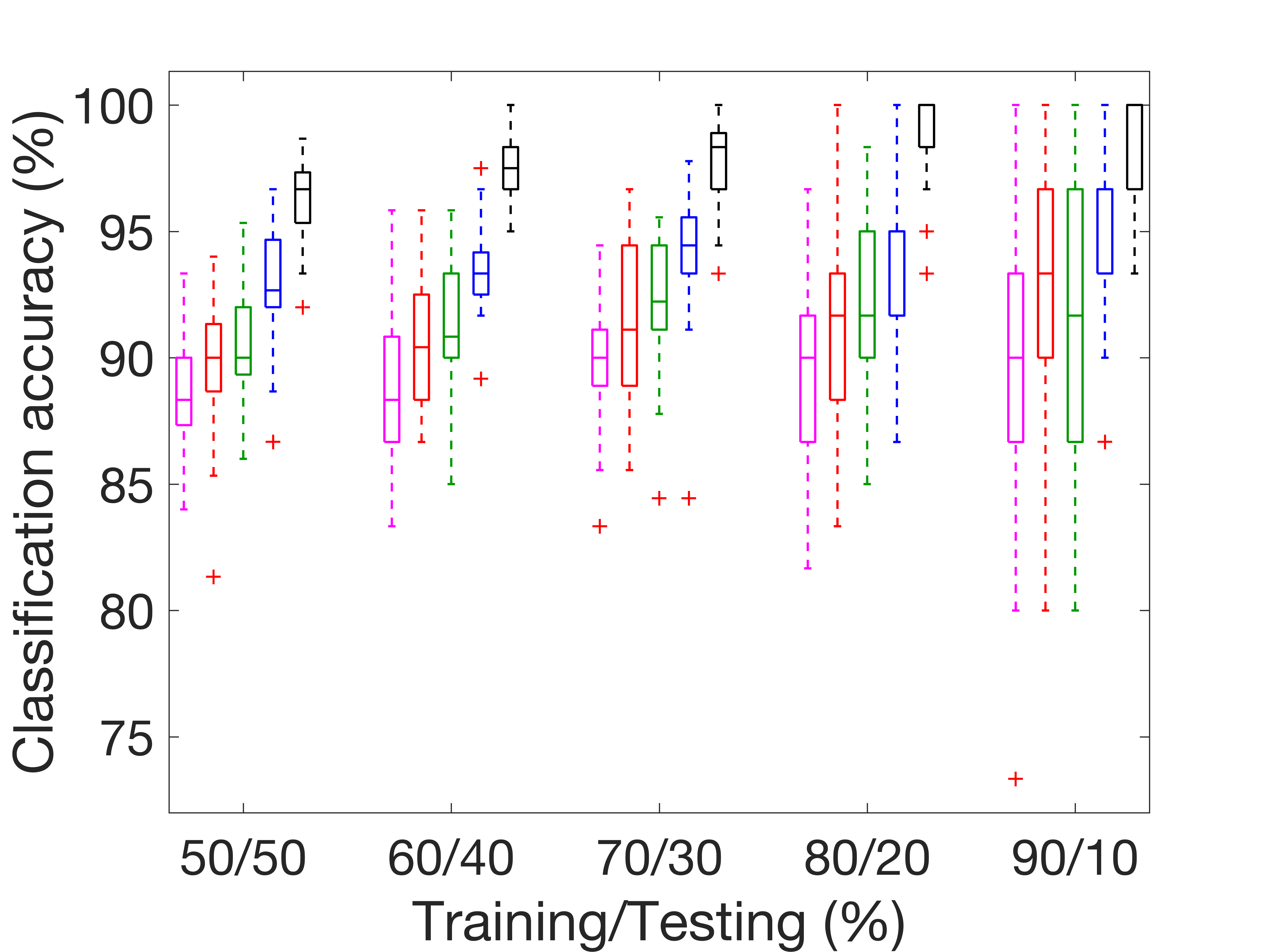}%
\label{fig_accuracy_syn}}
\hfil
\subfloat{\includegraphics[width=2.35in]{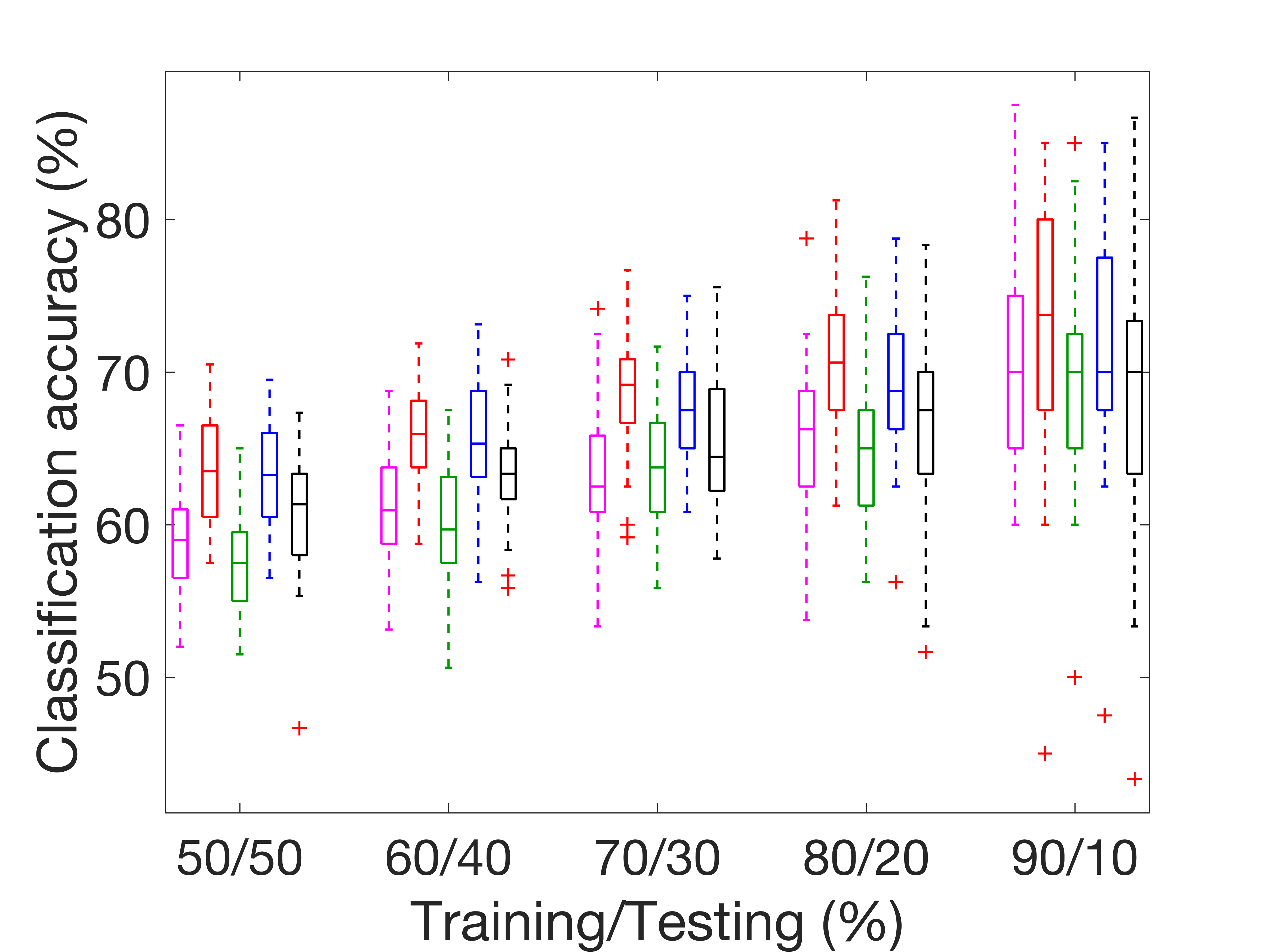}%
\label{fig_accuracy_real}}
\hfil
\subfloat{\includegraphics[width=2.35in]{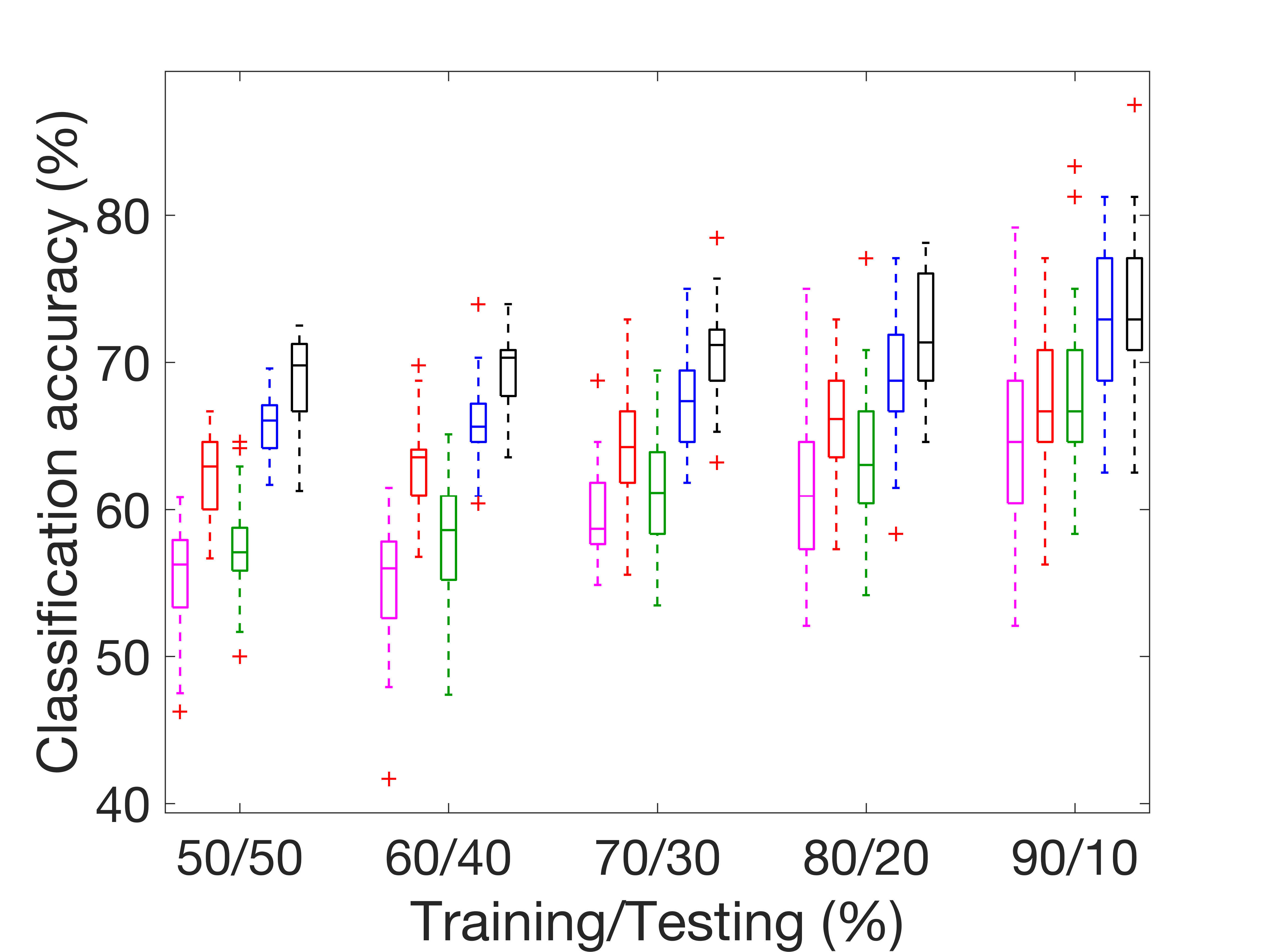}%
\label{fig_accuracy_outex}}\\
\addtocounter{subfigure}{-3}
\subfloat[SHREC'14 Synthetic]{\includegraphics[width=2.35in]{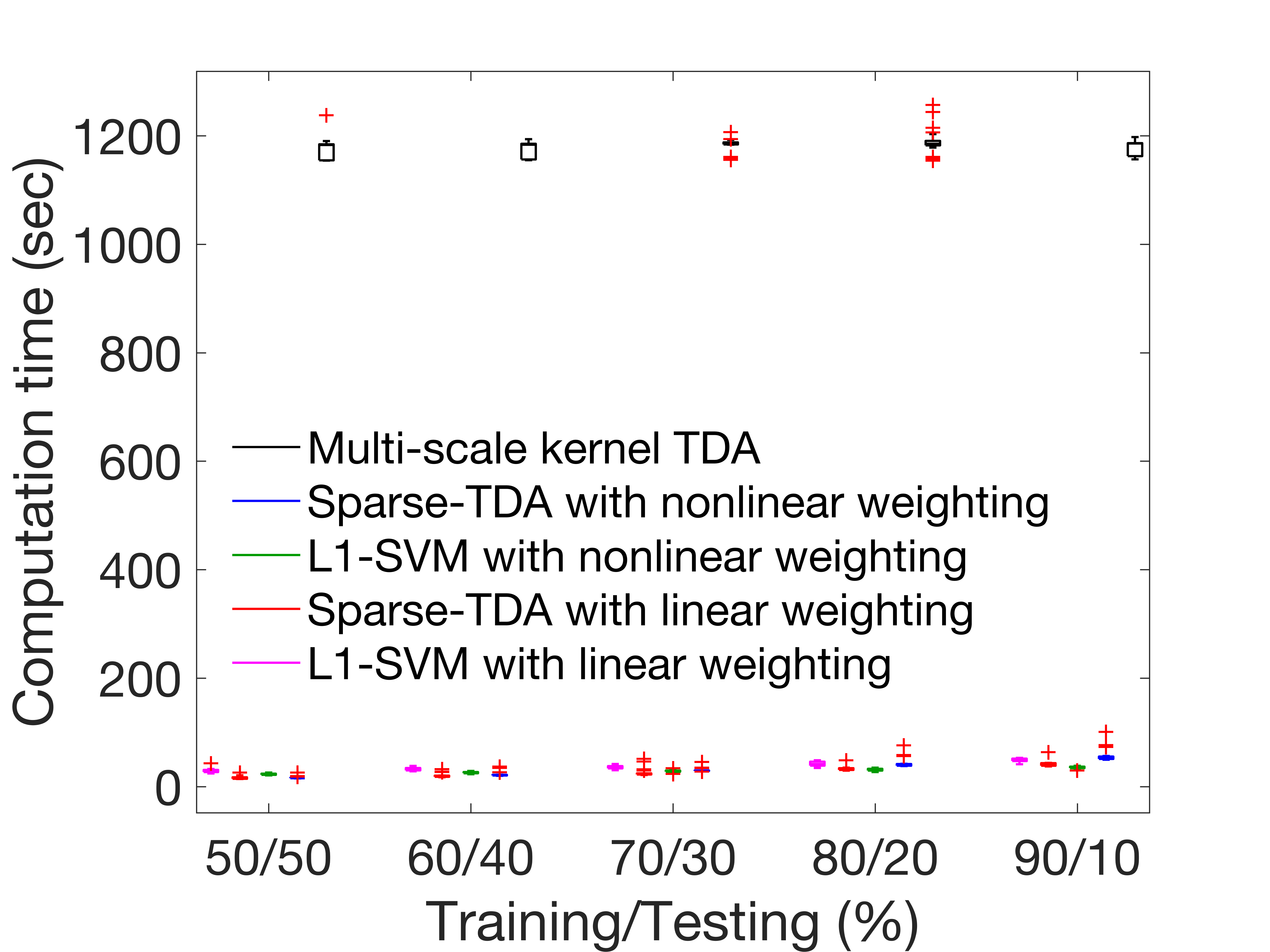}%
\label{fig_time_syn}}
\hfil
\subfloat[SHREC'14 Real]{\includegraphics[width=2.35in]{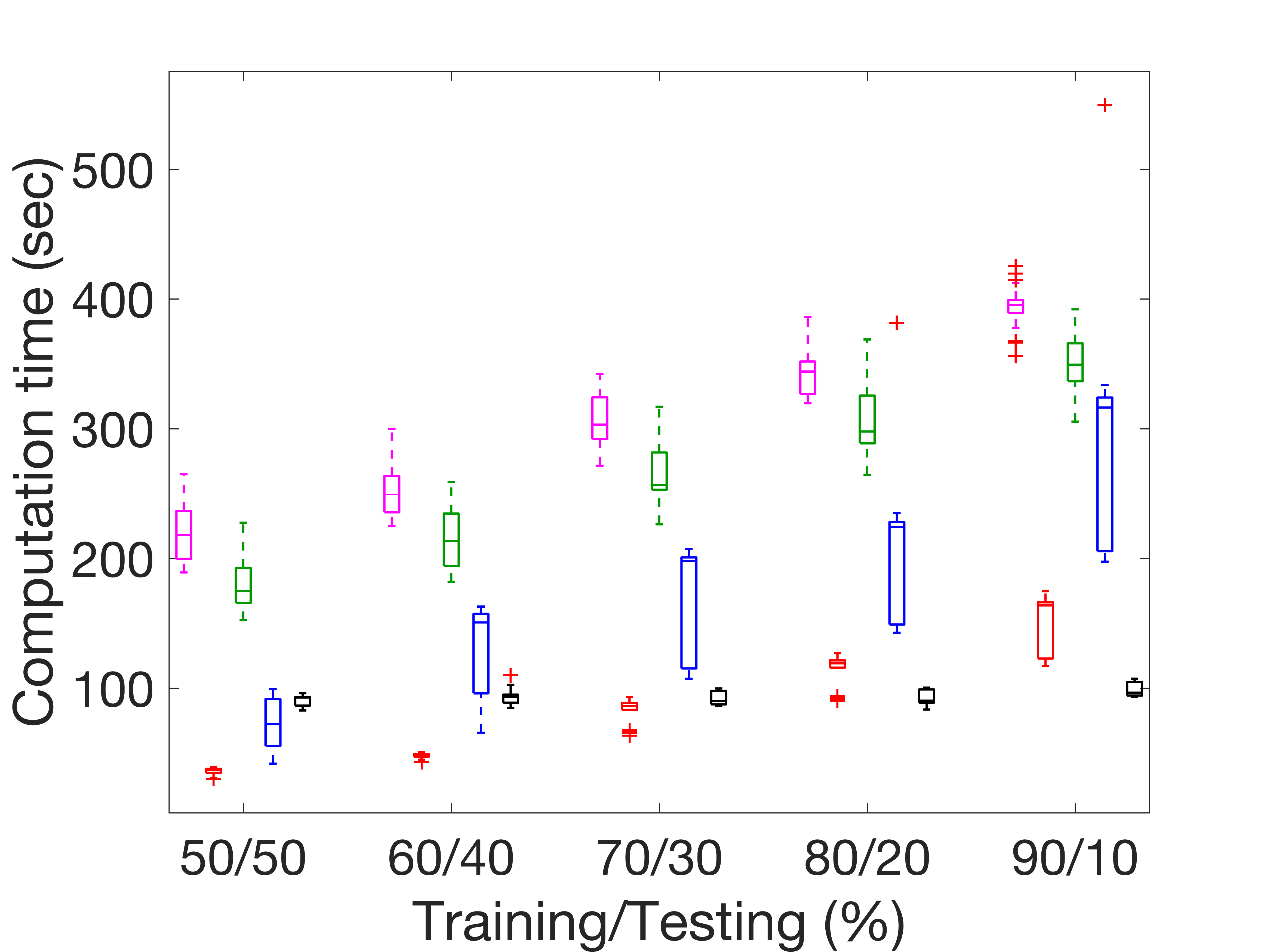}%
\label{fig_time_real}}
\hfil
\subfloat[OuTeX Texture]{\includegraphics[width=2.35in]{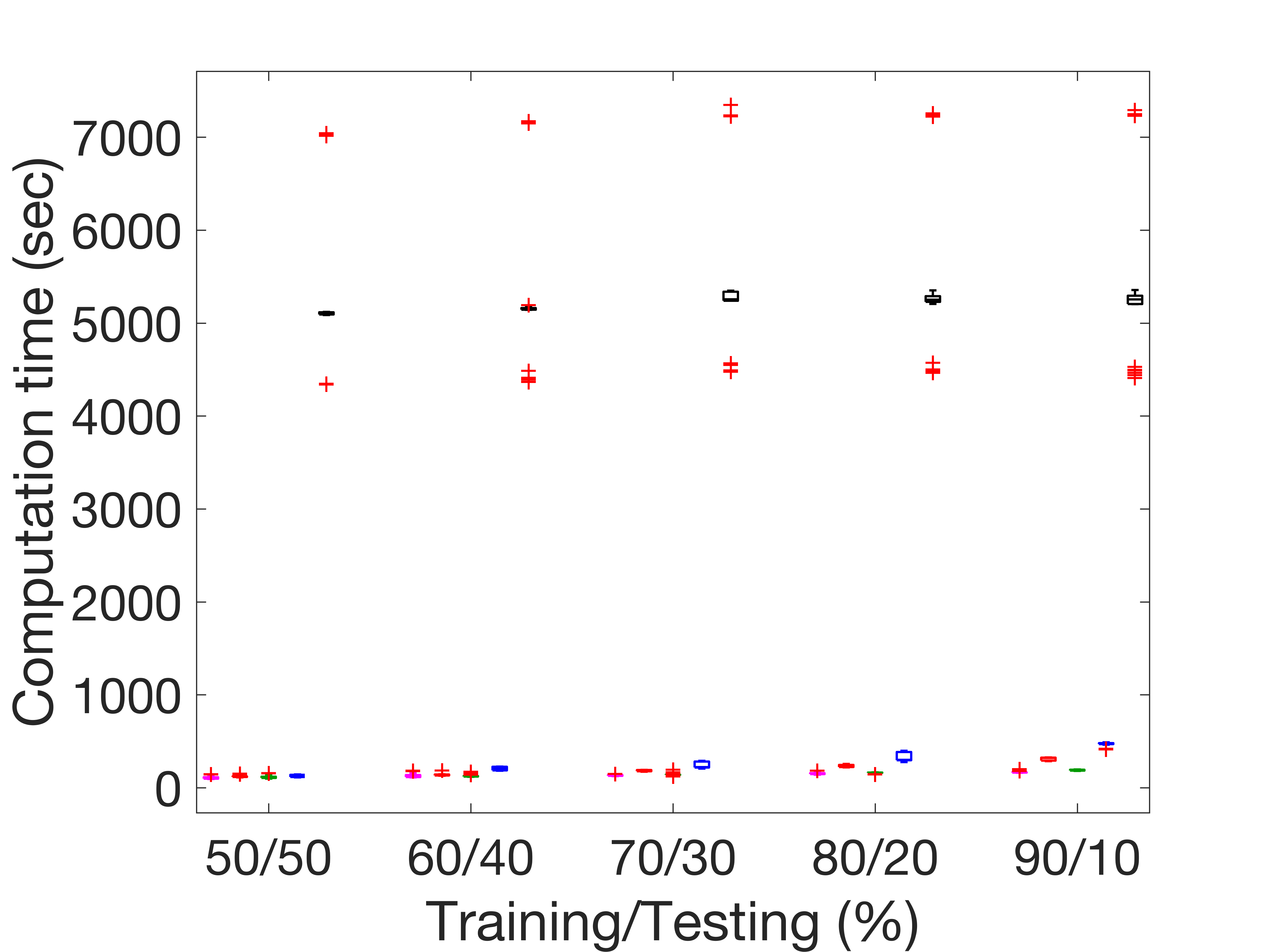}%
\label{fig_time_outex}}
\caption{Comparison of classification accuracy and training time among L1-SVM, Sparse-TDA, and state-of-the-art multi-scale kernel TDA methods for various training-testing partition ratios.}
\label{fig_acc_time}
\end{figure*}

\begin{table}[h]
\renewcommand{\arraystretch}{1.25}
\caption{Comparison of classifier training time (in s) with the same training and test data set split as in the original multi-scale kernel TDA article}
\centering
\begin{tabular}{|c|c|c|c|c|}
\hline
\multicolumn{2}{|c|}{\multirow{2}{*}{Method}} & SHREC'14 & SHREC'14 & OuTeX \\
\multicolumn{2}{|c|}{} & Synthetic & Real & Texture\\
\hline 
\multirow{2}{*}{L1-SVM} & LW & $35.4\pm 2.9$ & $305\pm 20.9$ & ${\bf 106\pm 14.8}$\\
\cline{2-5}
& NW & $28.5\pm 1.9$ & $267\pm 21.0$ & $113\pm 15.9$\\
\hline 
\multirow{2}{*}{Sparse-TDA} & LW & ${\bf 25.6\pm 7.0}$ & ${\bf 82.2\pm 9.3}$ & $120\pm 9.7$\\
\cline{2-5}
& NW & $30.5\pm 3.5$ & $171\pm 41.4$ & $131\pm 11.5$\\
\hline 
\multicolumn{2}{|c|}{Kernel TDA} & $1182 \pm 12.0 $& $92.3\pm 5.1$ & $5457\pm 979$ \\
\hline
\end{tabular}
\label{table_time}
\end{table}
Fig.~\ref{fig_acc_time} shows the trends in improving the classification accuracy and reducing the classifier training time, respectively, as a function of increasing training/testing split for all the benchmark data sets. Consistent with the results reported in Table~\ref{table_accuracy}, our classification accuracy is marginally inferior to that of the kernel TDA method for the SHREC'14 synthetic and OuTeX data sets. However, both our method variants marginally outperform the kernel method for the most challenging SHREC'14 real data set. The training time trends are also very similar to the results presented earlier in Table~\ref{table_time}, with more than an order of magnitude reduction for the SHREC'14 synthetic and the OuTeX data sets, and comparable values for the SHREC'14 real set. The increase in classifier training times with higher training/test splits is, however, slightly more for both our method variants as compared to the kernel TDA method due to the selection of more pixel samples as training size increases. Overall, we observe that for each of the benchmark data sets, at least one of our Sparse-TDA variants outperforms the kernel TDA method either in terms of classification accuracy or classifier training time. Moreover, Sparse-TDA outperforms both the L1-SVM variants in terms of classification accuracy for all the data sets, and achieves comparable computation time for the SHREC'14 synthetic and the OuTeX data sets, and a substantial reduction for the SHREC'14 real set.\looseness=-1

\section{Conclusions}
\label{sec:conclusions}
In this paper, we present a new method, referred as the Sparse-TDA algorithm, that provides a sparse realization of a TDA algorithm. More specifically, we combine optimized sparse sampling based on pivoted QR factorization with a state-of-the-art TDA method. Instead of persistence diagrams, we use a vector-based representation of persistent homology, called persistence images, with two different weighting functions to extract the topological features. 

The results are promising on three benchmark multi-way classification problems pertaining to 3D meshes of human posture recognition, both for real and synthetic shapes, and image texture detection. Our method gives similar classification accuracy and substantial reduction in training times as compared to a kernel TDA method that was earlier evaluated on these data sets. It also provides better accuracy and similar training times as compared to popular SVM classifiers. Such performance is, therefore, expected to lay the foundation for online adaptation of TDA on challenging data sets with a large number of classes in response to changes in the availability of training samples.        

In the future, we would like to further improve the accuracy of the Sparse-TDA method by designing our own weighting function for the persistence images. We would also like to come up with theoretical performance guarantees based on the characteristics of the data sets, particularly the training sample size for each individual class. Last but not the least, we plan to show the effectiveness of our method on other hard classification problems arising in robot visual perception and human face recognition.  
\ifCLASSOPTIONcompsoc
  \section*{Acknowledgments}
\else
  \section*{Acknowledgment}
\fi

We would like to thank The Boeing Company for sponsoring this work in part under contract \# SSOW-BRT-W0714-0004 and Dr. Tom Hogan for helpful discussions. The views and opinions expressed in the paper are, however, solely of the authors and do not necessarily reflect those of the sponsor. We also would like to thank the anonymous reviewers for their constructive comments.  \looseness=-1
 
\ifCLASSOPTIONcaptionsoff
  \newpage
\fi

%






\end{document}